\documentclass{sig-alternate-2013}
\usepackage{amssymb}
\usepackage{amsmath}
\usepackage{graphicx}
\usepackage[utf8]{inputenc}
\usepackage{color}
\usepackage{algorithm}
\usepackage[noend]{algorithmic}
\usepackage{wasysym} % needed for some symbols varhexagon hexagon pentagon
\usepackage{xstring}
\usepackage{graphicx}
\usepackage{float}
\usepackage{rotating}
\usepackage{lscape}
\usepackage{colortbl}
\usepackage[table]{xcolor}
\usepackage{url}

\usepackage{amssymb,amsmath}
\usepackage{graphicx}
\usepackage{xcolor}
\usepackage{float}
\usepackage{rotating}

\newcommand{\bbobdatapath}{ppdatamany/} % default output folder of rungeneric.py

\graphicspath{{\bbobdatapath}}

\def\RR{{\rm I\hspace{-0.50ex}R}}

\newcommand{\ma}[1]{\mathchoice{\mbox{\boldmath$\displaystyle#1$}}
  {\mbox{\boldmath$\textstyle#1$}} {\mbox{\boldmath$\scriptstyle#1$}}
  {\mbox{\boldmath$\scriptscriptstyle#1$}}}
\renewcommand{\ma}[1]{\mathnormal{\mathbf{#1}}}
\newcommand{\mstr}[1]{\mathrm{#1}}
\newcommand{\C}{ \ensuremath{\ma{C}} }
\newcommand{\A}{ \ensuremath{\ma{A}} }

\newcommand{\I}{ \ensuremath{\ma{I}} }

\newcommand{\dd}{n}
\newcommand{\R}{\mathbb{R}}
\newcommand{\vc}[1]{\textit{\textbf{#1}}}
\def\NormOI{{\mathcal N}  \hspace{-0.13em}\left({\ma{0}, \ensuremath{\ma{I}}\,}\right)}
\def\ONE{{\rm 1\hspace{-0.80ex}1}}
\def\Id{\ensuremath{\ma{I}}}

\def\x{\vc{x}}

\def\m{\vc{m}}

\usepackage{url}
\newfont{\mycrnotice}{ptmr8t at 7pt}
\newfont{\myconfname}{ptmri8t at 7pt}

\permission{Author's version}

\begin{document}

\title{A Computationally Efficient Limited Memory CMA-ES for Large Scale Optimization}% -- $Revision: 4512$}
% submission: revision 4512

 \numberofauthors{1}
 \author{
 \alignauthor
 Ilya Loshchilov \\
 \affaddr{Laboratory of Intelligent Systems}\\
 \affaddr{\'Ecole Polytechnique F\'ed\'erale de Lausanne, Switzerland}\\
 \email{ ilya.loshchilov@epfl.ch}
}
\maketitle
\begin{abstract}
We propose a computationally efficient limited memory Covariance Matrix Adaptation Evolution Strategy for large scale optimization, which we call the LM-CMA-ES. 
The LM-CMA-ES is a stochastic, derivative-free algorithm for numerical optimization of non-linear, non-convex optimization problems in continuous domain.
Inspired by the limited memory BFGS method of Liu and Nocedal (1989), 
the LM-CMA-ES samples candidate solutions according to a covariance matrix reproduced from $m$ direction vectors selected during the optimization process. The decomposition of the covariance matrix into Cholesky factors allows to reduce the time and memory complexity of the sampling to $O(mn)$, where $n$ is the number of decision variables. When $n$ is large (e.g., $n$ > 1000), even relatively small values of $m$ (e.g., $m=20,30$) are sufficient to efficiently solve fully non-separable problems and to reduce the overall run-time.
\end{abstract}

{ Problem Solving, Control Methods, and Search}



%%%%%%%%%%%%%%%%%%%%%%%%%%%%%%%%%%%%%
\section{Introduction}
%%%%%%%%%%%%%%%%%%%%%%%%%%%%%%%%%%%%%

The Covariance Matrix Adaptation Evolution Strategy (CMA-ES) is designed to learn dependencies between decision variables by adapting a covariance matrix which defines the sampling distribution of candidate solutions \cite{2003HansenCMA}. This algorithm constantly demonstrates good performance at various platforms for comparing continuous optimizers such as the Black-Box Optimization Benchmarking (BBOB) workshop \cite{2010HansenBBOBAllTables,2013LoshchilovHCMA} and the Special Session at Congress on Evolutionary Computation \cite{2009GarciaCEC2005results,2013LoshchilovCEC}. The CMA-ES was also extended to noisy \cite{hansen2009method}, expensive \cite{auger2013benchmarking,loshchilov2012self} and multi-objective optimization \cite{igel2007covariance}.

The principle advantage of the CMA-ES, the learning of dependencies between $n$ decision variables, also forms its main practical limitations such as $O(n^2)$ memory storage and $O(n^2)$ computational complexity per function evaluation \cite{ros2008simple}. These limitations may preclude the use of the CMA-ES for computationally cheap but large scale optimization problems (e.g., with $n>100$) if the internal computational cost of CMA-ES is greater than the cost of one function evaluation. On non-trivial large scale problems with $n>10000$ not only the internal computational cost of the CMA-ES becomes significant but it is becoming simply impossible to efficiently store the covariance matrix in memory. One may argue that there are very few known continuous domain real-world problems of that huge dimensionality. This situation probably will not change much before practitioners have a set of  tools that are able to efficiently search in such huge search spaces.

Several evolution strategies (ESs) have been proposed to deal with large scale optimization problems: $O(n)$ time and space complexity algorithms such as separable CMA-ES (sep-CMA-ES \cite{ros2008simple}) and linear time Natural Evolution Strategy (R1-NES \cite{sun2011linear}), L-CMA-ES \cite{knight2007reducing} with $O(m^2n)$ time and $O(mn)$ space complexity, where only $m$ dominant eigen-pairs of the covariance matrix are computed. The sep-CMA-ES learns only the scaling of variables. The R1-NES learns only the predominant eigen-direction. The L-CMA-ES learns $m$ dominant eigen-pairs, but its $O(m^2n)$ sampling complexity practically ends up with $O(n^2)$ when $m=\sqrt{n}$ as studied in \cite{knight2007reducing} for non-separable problems where multiple adaptation directions are required. 

The problem of growing time and space complexity when optimizing large scale problems is not new. It was addressed in gradient-based optimization community when it became clear that for $n>1000$ the storage of the approximate inverse Hessian matrix precludes the use of quasi-Newton methods such as Broyden–Fletcher\--Goldfarb–Shanno (BFGS) method \cite{1970ShannoBFGS}. As a solution, it was proposed not to store the matrix but to reconstruct it using information from the last $m$ iterations \cite{nocedal1980updating}. The final algorithm called the limited memory BFGS algorithm (L-BFGS or LM-BFGS) is still considered to be the state-of-the-art of large scale gradient-based optimization \cite{liu1989limited}.  In this paper, we demonstrate that a very similar idea can be used to reconstruct the covariance matrix in the CMA-ES to reduce the time and space complexity to $O(mn$).

The paper is organized as follows.
Section \ref{state} reviews Evolution Strategies (ESs) proposed for large scale optimization. 
 The LM-CMA-ES algorithm is described in section \ref{LMCMAalgo}. 
The experimental validation of LM-CMA-ES is reported and discussed in section \ref{expe}. Section \ref{conclusion} concludes the paper.
%and directions for further work are sketched in section .

%%%%%%%%%%%%%%%%%%%%%%%%%%%%%%%%%%%%%
\section{Evolution Strategies for Large Scale Optimization}
\label{state}
%%%%%%%%%%%%%%%%%%%%%%%%%%%%%%%%%%%%%

Historically, first Evolution Strategies \cite{1973RechenbergEvolutionsstrategie} were designed to perform the search without learning dependencies between variables which is a more recent development that gradually led to the CMA-ES algorithm \cite{1996HansenCMAES,2003HansenCMA}.
In this section, we discuss in detail the CMA-ES algorithm and its state-of-the-art derivatives for large scale optimization. For a recent comprehensible overview of Evolution Strategies, the interested reader is referred to \cite{2012HansenArnoldAugerES}. 

%%%%%%%%%%%%%%%%%%%%%%%%%%%%%%%%%%%%%
\subsection{The CMA-ES}
\label{sectionCMA}
%%%%%%%%%%%%%%%%%%%%%%%%%%%%%%%%%%%%%

The Covariance Matrix Adaptation Evolution Strategy \cite{1996HansenCMAES,2001HansenCMA,2003HansenCMA} is probably the most popular and in overall the most efficient Evolution Strategy. 

The ($\mu/\mu_{w},\lambda$)-CMA-ES is outlined in \textbf{Algorithm} \ref{CMAdefault}. 
At iteration $t$ of CMA-ES, a mean $\vc{m}^t$ of the mutation distribution (can be interpreted as an estimation of the optimum) is used to generate its $k$-th out of $\lambda$ candidate solution $\vc{x}_k \in \mathbb{R}^\dd$ 
	 (line \ref{CMAsampling}) by adding a random Gaussian mutation defined by a (positive definite) covariance matrix $\C^t \in \R^{n \times n}$ as
	
	\begin{equation}
  \vc{x}^t_k = {\mathcal N}  \hspace{-0.13em}\left({\vc{m}^t,{\sigma^t}^2 {\C}^t}\right) = \vc{m}^t + \sigma^t {\mathcal N}  \hspace{-0.13em}\left({\ma{0},{\C}^t}\right),
  \end{equation}
	
	where $\sigma^t$ is a mutation step-size. These $\lambda$ solutions then should be evaluated on an objective function $f$ (line \ref{CMAGenerateEnd}).
	The old mean of the mutation distribution is stored in $\vc{m}^{t}$ and a new mean $\vc{m}^{t+1}$ is computed as a \textit{weighted sum} 
	of the best $\mu$ parent individuals selected among $\lambda$ generated offspring individuals (line \ref{CMAComputeNewMean}).
	The weights $\vc{w}$ are used to control the impact of selected individuals, weights are usually higher for better ranked individuals (line \ref{CMAEScmaGiven}).
	
	The procedure of the adaptation of the step-size $\sigma^t$ in CMA-ES is inherited from the Cumulative Step-Size Adaptation Evolution Strategy (CSA-ES) \cite{1996HansenCMAES} and is controlled by evolution path $\vc{p}_{\sigma}^{t+1}$.
	Successful mutation steps $\frac{\vc{m}^{t+1}-\vc{m}^{t}}{\sigma^{t}}$ (line \ref{CMASigmaPathUpdate}) are tracked in the space of sampling, i.e., in the isotropic coordinate system defined by principal components of the covariance matrix $\C^t$. To update the evolution path $\vc{p}_{\sigma}^{t+1}$ a decay/relaxation factor $c_{\sigma}$ is used to decrease the importance of previously performed steps with time.
	The step-size update rule increases the step-size if the length of the evolution path $\vc{p}_{\sigma}^{t+1}$ is longer than 
	the expected length of the evolution path under random selection $\mathbb{E} \left\| \NormOI \right\|$,
  and decreases otherwise (line \ref{CMAStepSizeUpdate}). Expectation of $\left\| \NormOI \right\|$ is approximated by $\sqrt{n} (1 - \frac{1}{4 n} + \frac{1}{21 n^2} )$.
	A damping parameter $d_{\sigma}$ controls the change of the step-size.
	
	The covariance matrix update consists of two parts (line \ref{CMAupdate}): \textit{rank-one update} \cite{2001HansenCMA} and \textit{ rank-$\mu$ update} \cite{2003HansenCMA}.	The rank-one update computes evolution path $\vc{p}_c^{t+1}$ of successful moves of the mean $\frac{\vc{m}^{t+1}-\vc{m}^{t}}{\sigma^{t}}$ 
	of the mutation distribution in the given coordinate system (line \ref{CMAEvoPathUpdate}), 
	in a similar way as the evolution path $\vc{p}_{\sigma}^{t+1}$ of the step-size. 
	To stall the update of $\vc{p}_c^{t+1}$ when $\sigma$ increases rapidly, a $h_{\sigma}$ trigger is used (line \ref{CMAhsigma}).
	
	The rank-$\mu$ update computes a covariance matrix $\C_{\mu}$ as a weighted sum of covariances of successful steps of $\mu$ best individuals (line \ref{CMAplusCov}).
	The update of $\C$ itself is a replace of previously accumulated information by a new one with corresponding weights of importance (line \ref{CMAupdate}):
	$c_1$ for covariance matrix $\vc{p}^{t+1}_c {\vc{p}^{t+1}_c}^T$ of rank-one update and 
	$c_{\mu}$ for $\C_{\mu}$ of rank-$\mu$ update \cite{2003HansenCMA} such that $c_1 + c_{\mu} \leq 1$.
	Recently it was proposed to also take into account unsuccessful mutations in the \textit{"active" rank-$\mu$ update} \cite{2010HansenBBOBActiveCMA,2006ArnoldActiveCMA}. 

\begin{algorithm}[tb!]
\caption{The ($\mu/\mu_{w},\lambda$)-CMA-ES}
\label{CMAdefault}
\begin{algorithmic}[1]
\STATE{\textbf{given} $n \in \mathbb{N}_+$, $\lambda = 4 + \lfloor 3 \mstr{ln} \, n  \rfloor $, $\mu =  \lfloor \lambda/2   \rfloor $, 
											$\vc{w}_i = \frac{ \mstr{ln}(\mu + \frac{1}{2}) - \mstr{ln}\,i}{ \sum^{\mu}_{j=1}(\mstr{ln}(\mu + \frac{1}{2})-\mstr{ln}\,j)} \; \mstr{for} \; i=1 \ldots \mu$,
											$\mu_w = \frac{1}{\sum^{\mu}_{i=1} w^2_i}$,
											$c_{\sigma} = \frac{\mu_w + 2}{n+\mu_w+3}$,      $d_{\sigma} = 1 + c_{\sigma} +2 \, \mstr{max}(0,\sqrt{ \frac{\mu_w - 1}{n+1}}-1)$,
											$c_c = \frac{4}{n+4}$, $c_1 = \frac{2 \, \mstr{min}(1,\lambda/6)}{(n+1.3)^2 +\mu_w}$, $c_{\mu} = \frac{2 \, (\mu_w -2 + 1/{\mu_w})}{(n+2)^2+\mu_w}$} \label{CMAEScmaGiven}
\STATE{\textbf{initialize} $\vc{m}^{t=0} \in \R^{\dd}, \sigma^{t=0} > 0, \vc{p}^{t=0}_{\sigma} = \ma{0}, \vc{p}^{t=0}_{c} = \ma{0}, \C^{t=0} = \Id, t \leftarrow 0 $}
\REPEAT
  \FOR{$k = 1,\ldots,\lambda$} \label{CMAGenerateBegin}
			\STATE{ $\vc{x}_k = \vc{m}^t + \sigma^{t}  {{\mathcal N}  \hspace{-0.13em}\left({\ma{0},\C^{t}\,}\right)} $} \label{CMAsampling}
			\STATE{ $\vc{f}_k = f(\vc{x}_k)$} \label{CMAGenerateEnd}
  \ENDFOR
	\STATE{ $ \vc{m}^{t+1} \leftarrow \sum_{i=1}^{\mu} \vc{w}_i \vc{x}_{i:\lambda} \;$} // the symbol $i:\lambda$ denotes $i$-th best individual on $f$ \label{CMAComputeNewMean}
	\STATE{ $ \vc{p}^{t+1}_{\sigma} \leftarrow (1 - c_{\sigma}) \vc{p}^{t}_{\sigma} + \sqrt{c_{\sigma}(2-c_{\sigma})} \sqrt{\mu_w} {\C^t}^{-\frac{1}{2}} \frac{\vc{m}^{t+1}-\vc{m}^{t}}{\sigma^{t}} $} \label{CMASigmaPathUpdate}
	\STATE{ $ h_{\sigma} = \ONE_{ \left\| p^{t+1}_{\sigma} \right\| < \sqrt{1 - (1-c_{\sigma})^{2(t+1)}}(1.4 + \frac{2}{n+1}) \, \mathbb{E} \left\| \NormOI \right\|  } $} \label{CMAhsigma}
	\STATE{ $ \vc{p}^{t+1}_{c} \leftarrow (1 - c_{c}) \vc{p}^{t}_{c} + h_{\sigma} \sqrt{c_{c}(2-c_{c})} \sqrt{\mu_w} \frac{\vc{m}^{t+1}-\vc{m}^{t}}{\sigma^{t}} $} \label{CMAEvoPathUpdate}
	\STATE{ $ \C_{\mu} = \sum^{\mu}_{i=1} w_i \frac{\x_{i:\lambda} - \m^t}{\sigma^t} \times \frac{(\x_{i:\lambda} - \m^t)^T}{\sigma^t}$ }  \label{CMAplusCov}
	\STATE{ $ \C^{t+1} = (1 - c_1 - c_{\mu}) \C^t	+ 
						c_1 \underbrace{\vc{p}^{t+1}_c {\vc{p}^{t+1}_c}^T}_{\mstr{\tiny rank-one\,update}} + 
						c_{\mu} \hspace{-1.9em} \underbrace{\C^{+}_{\mu}}_{\mstr{rank-\mu \,update}}$}
						\label{CMAupdate}
	\STATE{ $ \sigma^{t+1} \leftarrow \sigma^{t} \mstr{exp}
	          ( \frac{c_{\sigma}}{d_{\sigma}} (  \frac{\left\| \vc{p}^{t+1}_{\sigma} \right\|}{ \mathbb{E} \left\| \NormOI \right\| } - 1  )) $} \label{CMAStepSizeUpdate}
  \STATE{ $ t \leftarrow t + 1$}
\UNTIL{ \textit{stopping criterion is met} }
\end{algorithmic}
\end{algorithm}

  In CMA-ES, the factorization of the covariance $\C$ into $\A \A^T=\C$ is needed to sample the multivariate normal distribution (line \ref{CMAsampling}). The eigendecomposition with $O(n^3)$ complexity is used for the factorization. Already in the original CMA-ES it was proposed to perform the eigendecomposition every $n/10$ generations (not shown in \textbf{Algorithm} \ref{CMAdefault}) to reduce the complexity per function evaluation to $O(n^2)$ 
																														
\subsection{Large Scale Variants}\label{largescale}

The original CMA-ES has $O(n^2)$ time and space complexity that precludes its applications for large scale optimization with $n\gg100$. To enable the algorithm for large scale optimization, a linear time and space version called sep-CMA-ES was proposed in \cite{ros2008simple}. The algorithm does not learn dependencies but the scaling of variables by restraining the covariance matrix update to the diagonal elements: %of the matrix:

	\begin{eqnarray}
	\lefteqn{c^{t+1}_{jj} = (1 - c_{cov}) c^t_{jj} + \frac{1}{\mu_{cov}}\left(\vc{p}_c^{t+1}\right)^2_j + } \nonumber \\
	&& c_{ccov}\left(1-\frac{1}{\mu_{ccov}}\right) \sum_{i=1}^{\mu} w_i c^t_{jj} \left({z_{i:\lambda}}^{t+1}\right)^2_j, j=1,\ldots,n
	\end{eqnarray}
 where, for $j=1,\ldots,n$ the $c_{jj}$ are the diagonal elements of $\C^t$ and the $\left({z_{i:\lambda}}^{t+1}\right)_j = \left({x_{i:\lambda}}^{t+1}\right)_j/(\sigma^t \sqrt{(c_{jj})}$.

This update reduces the computational complexity to $O(n)$ and allows to exploit problem separability, thus the original property of being rotationally invariant is lost. The algorithm demonstrated good performance on separable problems and even outperformed CMA-ES on non-separable Rosenbrock function for $n>100$.

A novel Natural Evolution Strategy (NES) variant, the Rank-One NES (R1-NES), which uses a low rank approximation of the search distribution covariance matrix was proposed recently by \cite{sun2011linear}. The algorithm adapts the search distribution according to the natural gradient with a particular parametrization of the covariance matrix,

	\begin{equation}
  \C = \sigma^2 (\I + \vc{u}\vc{u}^T),
  \end{equation}

where $u$ and $\sigma$ are the parameters to be adjusted. The adaptation of the predominant eigen-direction $\vc{u}$ allows the algorithm to solve highly non-separable problems while maintaining only $O(n)$ time and space complexity.

A version of CMA-ES with a limited memory storage also called limited memory CMA-ES (L-CMA-ES) was proposed by \cite{knight2007reducing}. The L-CMA-ES uses the
$m$ eigen-vectors and eigen-values spanning the $m$-dimensional
dominant subspace of the $n$-dimensional covariance matrix \C. 
 The authors adapted a singular value decomposition updating 
 algorithm developed in \cite{brand2006fast} that allowed to avoid 
the explicit computation and storage of the covariance matrix. 
For $m < n$ the performance in terms of number of function evaluations gradually decreases while enabling the search in $\R^n$ for $n>10000$. However, the computational complexity of $O(m^2n)$ practically (for $m$ in order of $\sqrt{n}$ \cite{knight2007reducing}) leads to the same limitations as for the original CMA-ES.

The ($\mu/\mu_{w},\lambda$)-Cholesky-CMA-ES proposed in \cite{2009Suttorp11CMA} is of special interest in this paper because the LM-CMA-ES is based on this algorithm. The Cholesky-CMA represents a version of CMA-ES with rank-one update where instead of performing the factorization of the covariance matrix $\C^t$ into $\A^t {\A^t}^T=\C^t$, the Cholesky factor $\A^t$ and its inverse ${\A^t}^{-1}$ are iteratively updated. From \textbf{Theorem 1} \cite{2009Suttorp11CMA} it follows that if $\C^t$ is updated as 

	\begin{equation}
  \C^{t+1} = \alpha \C^t + \beta \vc{v}^t {\vc{v}^t}^T,
  \end{equation}
	
	where $\vc{v} \in \R^n$ is given in the decomposition form $\vc{v}^t = \A^t \vc{z}^t$, and $\alpha,\beta \in \R^+$, then for $\vc{z} \neq \vc{0}$ a Cholesky factor of the matrix $\C^{t+1}$ can be computed by
	
	\begin{equation} 
	\label{Aupdate}
  \A^{t+1} = \sqrt{\alpha} \A^t + \frac{\sqrt{\alpha}}{{\left\| \vc{z}^t \right\|}^2} \left( \sqrt{1 + \frac{\beta}{\alpha} {{\left\| \vc{z}^t \right\|}^2}} -1 \right) [\A^t \vc{z}^t] {\vc{z}^t}^T,
  \end{equation}
	
	for $\vc{z}_t=\vc{0}$ we have $\A^{t+1}=\sqrt{\alpha}\A^t$. From the \textbf{Theorem 2} \cite{2009Suttorp11CMA} it follows that if ${\A^{-1}}^{t}$ is the inverse of $\A^t$, then the inverse of $\A^{t+1}$ can be computed by

	\begin{equation}
	\label{Ainvupdate}
  {\A^{-1}}^{t+1} = \frac{1}{\sqrt{\alpha}} {\A^{-1}}^{t} - \frac{1}{\sqrt{\alpha}{\left\| \vc{z}^t \right\|}^2} \left( 1- \frac{1}{\sqrt{1 + \frac{\beta}{\alpha} {{\left\| \vc{z}^t \right\|}^2}}} \right) \vc{z}^t [{\vc{z}^t}^T {\A^{-1}}^{t}],
  \end{equation}
	
	for $\vc{z}^t \neq \vc{0}$ and by ${\A^{-1}}^{t+1}=\frac{1}{\sqrt{\alpha}} {\A^{-1}}^{t}$ for $\vc{z}^t = \vc{0}$.

\begin{algorithm}[tb!]
\caption{The ($\mu/\mu_{w},\lambda$)-Cholesky-CMA-ES}
\label{CholCMA}
\begin{algorithmic}[1]
\STATE{\textbf{given} $n \in \mathbb{N}_+$, $\lambda = 4 + \lfloor 3 \mstr{ln} \, n  \rfloor $, $\mu =  \lfloor \lambda/2   \rfloor $, 
											$w_i = \frac{ \mstr{ln}(\mu + 1) - \mstr{ln}(i)}{\mu \mstr{ln}(\mu + 1) - \sum_{j=1}^{\mu} \mstr{ln}(j)}; i=1 \ldots \mu$,
											$\mu_w = \frac{1}{\sum^{\mu}_{i=1} w^2_i}$,
											$c_{\sigma} = \frac{\sqrt{\mu_{w}}}{
											\sqrt{n} + \sqrt{\mu_{w}}}$,      
											$d_{\sigma} = 1 + c_{\sigma} +2 \, \mstr{max}(0,\sqrt{ \frac{\mu_w - 1}{n+1}}-1)$,
											$c_c = \frac{4}{n+4}$, $c_1 = \frac{2}{ {(n + \sqrt{2})}^2}$
											} \label{CholCMAEScmaGiven}
\STATE{\textbf{initialize} $\vc{m}^{t=0} \in \R^{\dd}, \sigma^{t=0} > 0, \vc{p}^{t=0}_{\sigma} = \ma{0}, \vc{p}^{t=0}_{c} = \ma{0}, \A^{t=0} = \Id, \A^{t=0}_{inv} = \Id, t \leftarrow 0 $}
\REPEAT
  \FOR{$k = 1,\ldots,\lambda$} \label{CholCMAGenerateBegin}
		\STATE{ $\vc{z}_k = {{\mathcal N}  \hspace{-0.13em}\left({\ma{0},\I}\right)} $} \label{CholCMAsampling1}
			\STATE{ $\vc{x}_k = \vc{m}^t + \sigma^{t} \A \vc{z}_k $} \label{CholCMAsampling2}
			\STATE{ $\vc{f}_k = f(\vc{x}_k)$} \label{CholCMAGenerateEnd}
  \ENDFOR
	\STATE{ $ \vc{m}^{t+1} \leftarrow \sum_{i=1}^{\mu} \vc{w}_i \vc{x}_{i:\lambda} \;$}  \label{CholCMAComputeNewMean}
		\STATE{ $ \vc{z}_w \leftarrow \sum_{i=1}^{\mu} \vc{w}_i \vc{z}_{i:\lambda} \;$} \label{CholCMAComputeNewZMean}
	\STATE{ $ \vc{p}^{t+1}_{\sigma} \leftarrow (1 - c_{\sigma}) \vc{p}^{t}_{\sigma} + \sqrt{c_{\sigma}(2-c_{\sigma})} \sqrt{\mu_w} \vc{z}_w$} \label{CholCMASigmaPathUpdate}
	\STATE{ $ \vc{p}^{t+1}_{c} \leftarrow (1 - c_{c}) \vc{p}^{t}_{c} + \sqrt{c_{c}(2-c_{c})} \sqrt{\mu_w} \A \vc{z}_w$} \label{CholCMAEvoPathUpdate}
	\STATE{$ \vc{v} \leftarrow \A_{inv}^t \vc{p}_c  $}
	\STATE{$ \A^{t+1} = \sqrt{1 - c_1} \A^t + \frac{\sqrt{1 - c_1}}{{\left\| \vc{v}^t \right\|}^2} \left( \sqrt{1 + \frac{c_1}{1 - c_1} {{\left\| \vc{v}^t \right\|}^2}} -1 \right) \vc{p}_c {\vc{v}^t}^T $} \label{CholAUpdate}
	\STATE{$ {\A_{inv}^{t+1}} = \frac{1}{\sqrt{1 - c_1}} {\A_{inv}^t} - 
	\frac{1}{\sqrt{1 - c_1}{\left\| \vc{v}^t \right\|}^2} 
	\left( 1- \frac{1}{\sqrt{1 + \frac{c_1}{1 - c_1} {{\left\| \vc{v}^t \right\|}^2}}} \right) 
	{\vc{v}}^t [{{\vc{v}}^t}^T {\A_{inv}^t}], 
	$}						\label{CholAInvUpdate}
	\STATE{ $ \sigma^{t+1} \leftarrow \sigma^{t} \mstr{exp}
	          ( \frac{c_{\sigma}}{d_{\sigma}} (  \frac{\left\| \vc{p}^{t+1}_{\sigma} \right\|}{ \mathbb{E} \left\| \NormOI \right\| } - 1  )) $} \label{CholCMAStepSizeUpdate}
  \STATE{ $ t \leftarrow t + 1$}
\UNTIL{ \textit{stopping criterion is met} }
\end{algorithmic}
\end{algorithm}

The ($\mu/\mu_{w},\lambda$)-Cholesky-CMA-ES is outlined in \textbf{Algorithm} \ref{CholCMA}. 
As well as in the original CMA-ES, Cholesky-CMA-ES proceeds by sampling $\lambda$ candidate solutions (lines \ref{CholCMAGenerateBegin} - \ref{CholCMAGenerateEnd})  and taking into account the most successful $\mu$ out of $\lambda$ solutions in the evolution paths adaptation (lines \ref{CholCMASigmaPathUpdate} and 
\ref{CholCMAEvoPathUpdate}). However, the eigen-decomposition procedure is not required anymore because the Cholesky factor and its inverse are updated incrementally (line \ref{CholAUpdate} and \ref{CholAInvUpdate}). This simplifies a lot the implementation of the algorithm and reduces its time complexity to $O(n^2)$. A postponed update of the Cholesky factors every $O(n)$ iterations would not reduce the asymptotic  complexity further (as it does in the original CMA-ES) because the quadratic complexity will remain due to matrix-vector multiplications needed to sample new individuals. 

The non-elitist Cholesky-CMA is a good alternative to the original CMA-ES and demonstrates a comparable performance \cite{2009Suttorp11CMA}. While it has the same computational and memory complexity, the lack of rank-$\mu$ update may deteriorate its performance on problems where it is essential. 

\section{The LM-CMA-ES}\label{LMCMAalgo}

In this section, we first present main components of the computationally cheap limited memory CMA-ES and then introduce the algorithm itself. 
 The components are: a procedure for reconstruction of Cholesky factors of a covariance matrix using stored direction vectors, a procedure to store these vectors and a new procedure for step-size adaptation.

\subsection{Reconstruction of Cholesky factors}

The idea to reconstruct the inverse Hessian matrix in the BFGS method \cite{nocedal1980updating} enabled its application for large scale gradient-based optimization. While the CMA-ES is a gradient-free algorithm, the two algorithms are indeed similar with a difference that the latter estimates the gradient in a stochastic way. This observation inspired us to investigate whether a similar matrix reconstruction procedure can be used in CMA-ES as well to reduce its time and space complexity.

As can be seen, the only use of Cholesky factor $\A^t$ in \textbf{Algorithm} \ref{CholCMA} is for sampling of new solutions after $\A^t \vc{z}_k$ or for its own update to $\A^{t+1}$. By setting $a=\sqrt{1-c_1}$ and $b^t=\frac{\sqrt{1-c_1}}{{\left\| \vc{v}^t \right\|}^2} \left( \sqrt{1 + \frac{c_1}{1-c_1} {{\left\| \vc{v}^t \right\|}^2}} -1 \right)$, one can rewrite the line (\ref{CholAUpdate}) as 

	\begin{equation} 
	\label{AupdateRew}
  \A^{t+1} = a \A^t + b^t \vc{p}_c^t {\vc{v}^t}^T,
  \end{equation}

	In the following, we show how the vectors needed to sample new candidate solutions can be obtained without an explicit storage of Cholesky factors. At iteration $t=0$, $\A^0=\I$ and $\A^0 \vc{z}=\vc{z}$ in line (\ref{CholCMAsampling2}) of \textbf{Algorithm} \ref{CholCMA}, the new updated Cholesky factor $\A^{1}=a \I + b^{0} \vc{p}_c^{0} {\vc{v}^{0}}^T$. At iteration $t=1$, $\A^{1} \vc{z} = (a \I + b^{0} \vc{p}_c^{0} {\vc{v}^{0}}^T) \vc{z} = a \vc{z} + b^0 \vc{p}_c^0 ({\vc{v}^0}^T \vc{z})$ and $\A^2=a (a \I + b^{0} \vc{p}_c^{0} {\vc{v}^{0}}^T) + b^1 \vc{p}_c^1 {\vc{v}^1}^T$. Thus, a very simple iterative procedure which scales as $O(mn)$ can be used to sample candidate solutions in $\RR^n$  according to the Cholesky factor $\A^t$ reconstructed from $m$ pairs of vectors $\vc{p}_c^t$ and $\vc{v}^t$.

\begin{algorithm}[tb!]
\caption{ Az(): Cholesky factor - vector update}
\label{CholVector}
\begin{algorithmic}[1]
\STATE{\textbf{given} $\vc{z} \in \R^n,  m \in \mathbb{Z}_+, \vc{j} \in \mathbb{Z}^m_+, \vc{P} \in \R^{m \times n}, \vc{V} \in \R^{m \times n}, \vc{b} \in \R^m, a \in [0,1]$
											} \label{CholVectorGiven1}
\STATE{\textbf{initialize} $\vc{x}  \leftarrow \vc{z}$}
  \FOR{$t = 1,\ldots,min(m,\left|\vc{j}\right|)$} \label{CholVectorGenerateBegin}
		\STATE{ $k \leftarrow \vc{b}_{\vc{j}_t} \vc{V}_{(\vc{j}_t,:)} \cdot \vc{x} $} \label{CholVector1} 
		\STATE{ $\vc{x} \leftarrow  a \vc{x} + k \vc{P}_{(\vc{j}_t,:)}$} \label{CholVector3}
  \ENDFOR
\STATE{ \textbf{return} $\vc{x}$ }
\end{algorithmic}
\end{algorithm}

	\begin{algorithm}[tb!]
\caption{ Ainvz(): inverse Cholesky factor - vector update}
\label{InvCholVector}
\begin{algorithmic}[1]
\STATE{\textbf{given} $\vc{z} \in \R^n, m \in \mathbb{Z}_+, \vc{j} \in \mathbb{Z}^m, \vc{V} \in \R^{m \times n}, \vc{d} \in \R^m, c \in [0,1]$
											} \label{InvCholVectorGiven}
\STATE{\textbf{initialize} $\vc{x}  \leftarrow \vc{z}$}
  \FOR{$t = 1,\ldots,min(m,\left|\vc{j}\right|)$} \label{InvCholVectorGenerateBegin2}
		\STATE{ $k \leftarrow \vc{d}_{\vc{j}_t}  \vc{V}_{(\vc{j}_t,:)} \cdot \vc{x} $} \label{InvCholVector1} 
		\STATE{ $\vc{x} \leftarrow  c \vc{x} - k \vc{V}_{(\vc{j}_t,:)}$} \label{InvCholVector3}
  \ENDFOR
\STATE{ \textbf{return} $\vc{x}$ }
\end{algorithmic}
\end{algorithm}

\begin{algorithm}[tb!]
\caption{ UpdateSet(): direction vectors selection }
\label{Selection}
\begin{algorithmic}[1]
\STATE{\textbf{given} $m \in \R^+, \vc{j} \in \mathbb{Z}_+^m, \vc{l} \in \mathbb{Z}_+^m, t \in \mathbb{Z}_+, N_{steps} \in \mathbb{Z}_+$
											} \label{sel1}
	\IF{$ t < m$}
		\STATE{$ \vc{j}_t \leftarrow t$}	\label{sel2}
	\ELSE
		\STATE{$ i_{min} \leftarrow  1+argmin_i\left(\vc{l}_{\vc{j}_{i+1}}-\vc{l}_{\vc{j}_i}\right),|1\leq i \leq (m-1)$}	\label{sel3}
		\IF{$ \vc{l}_{\vc{j}_{i_{min}}}-\vc{l}_{\vc{j}_{i_{min}-1}} \geq N_{steps}$}	\label{sel4}
			\STATE{$ i_{min} \leftarrow 1$}	\label{sel5}
		\ENDIF{}
		\IF{$ i_{min} \neq m$}	\label{sel6}
			\STATE{$ \vc{j}_{tmp} \leftarrow \vc{j}_{i_{min}}$}	\label{sel7}
			\FOR{$i = i_{min},\ldots,m-1$} \label{sel8}
				\STATE{ $\vc{j}_i \leftarrow \vc{j}_{i+1}$} \label{sel9} 
			\ENDFOR
			\STATE{ $\vc{j}_m \leftarrow \vc{j}_{tmp}$} \label{sel10} 
		\ENDIF{}
	\ENDIF{}
	\STATE{$j_{cur} \leftarrow \vc{j}_{min(t + 1, m)}$}	\label{sel12}
	\STATE{$\vc{l}_{j_{cur}} \leftarrow t$}	\label{sel13}
\STATE{ \textbf{return}: $\vc{j}_{cur}$, $\vc{j}$, $\vc{l}$  }	\label{sel34}
\end{algorithmic}
\end{algorithm}	
	
	The pseudo-code of the procedure to reconstruct $\vc{x}=\A^t \vc{z}$ from $m$ direction vectors\footnote{more precisely, we mean $m$ evolution paths $\vc{p}_c$ and their inverses $\vc{v}$ but for brevity we say $m$ direction vectors} is given in \textbf{Algorithm} \ref{CholVector}. At each iteration of reconstruction of $\vc{x}=\A^t \vc{z}$  (lines \ref{CholVectorGenerateBegin} - \ref{CholVector1}), $\vc{x}$ is updated as a sum of $a$-weighted version of itself and ${b}^t$-weighted evolution path $\vc{p}_c^t$ scaled by the dot product of $\vc{v}^t$ and $\vc{x}$. As can be seen, the \textbf{Algorithm} uses $\vc{j}(t)$ indexation instead of $t$. This is simply a convenient way to have references to matrices $\vc{P}$ and $\vc{V}$ which store $\vc{p}_c^t$ and $\vc{v}^t$ vectors, respectively. In the next subsection, we will show how to efficiently manipulate these vectors.
	
	A very similar approach can be used to reconstruct $\vc{x}={\A^t}^{-1} \vc{z}$, for the sake of reproducibility the pseudo-code is given in \textbf{Algorithm} \ref{InvCholVector} for $c=1/\sqrt{1-c_1}$ and $d^t= \frac{1}{\sqrt{1-c_1}{\left\| \vc{v}^t \right\|}^2} \times \\ \times \left( 1- \frac{1}{\sqrt{1 + \frac{c_1}{1-c_1} {{\left\| \vc{v}^t \right\|}^2}}} \right)$. The computational complexity of both procedures scales as $O(mn)$.
	
\subsection{Direction Vectors Selection and Storage}	

It is an open question how to use only $m\ll n$ direction vectors to obtain a comparable amount of useful information as stored in the covariance matrix of the original CMA-ES. For large $n$ and $\lambda\ll n$, evolution path vectors $\vc{p}^t_c$ from the last $m$ iterations are likely to be quite similar and therefore to contain only some local information. 

In this paper, we propose a simple approach which forces $m$ selected vectors to be at approximately the same distance from each other in terms of number of iterations, but at most with the distance of $N_{steps}$ from each other given that the $m$-th vector is the one from the last iteration. This selection procedure is outlined in \textbf{Algorithm} \ref{Selection} which outputs an array of pointers $\vc{j}$ such that $\vc{j}_1$ points out to a row in matrices $\vc{P}$ and $\vc{V}$ with the oldest saved vectors $\vc{p}_c$ and $\vc{v}$ which will be taken into account during the reconstruction procedure. The higher the index $i$ of $\vc{j}_i$ the more recent the corresponding direction vector is. The index $j_{cur}$ points out to the oldest vector which will be replaced by the newest one in the same iteration when the procedure is called. The rule to choose a vector to be replaced is the following: find a pair of consecutively saved vectors with the closest distance (in terms of number of iterations, stored in $\vc{l}$) between each other (line \ref{sel3}), if this distance is smaller than $N_{steps}$ then the most recent vector will be removed by assigning $j_{cur} \leftarrow i_{min}$, otherwise the oldest vector among $m$ saved vectors should be removed.
 Thus, the procedure gradually replaces vectors in a way to keep them at approximately the same distance, but at most at distance of $N_{steps}$ iterations.

\subsection{Population Success Rule}	

An elegant success rule for step-size adaptation called the \textit{median success rule} was recently proposed in \cite{elhara2013median}. It is applicable to non-elitist multi-recombinant evolution strategies. The median success rule compares the median fitness of the population to a fitness from the previous iteration. The comparison fitness is chosen to achieve a target
success rate of 1/2. The empirical validation demonstrated that the median success rule is competitive to CSA \cite{elhara2013median}.

In practice, one should count the number $K_{succ}$ of individuals in the current population better than some $j$-th best individual of the previous population, where $j$ depends on $n$ and $\lambda$ but can be set to be $0.3\lambda$ \cite{elhara2013median}. Then, a normalized measurement 

	\begin{equation} 
	\label{Rew}
  z \leftarrow \frac{2}{\lambda} \left( K_{succ} - \frac{\lambda+1}{2}  \right)
  \end{equation}
	can be computed such that $z\geq 0$ iff the median individual was successful.
	
	The step-size is adapted as 
	
	\begin{equation} 
	\label{MruleSigma}
  \sigma \leftarrow \sigma \exp\left(\frac{s}{d_{\sigma}}\right),
  \end{equation}
	
	where $s \leftarrow (1 - c_{\sigma}) s + c_{\sigma} z$ and $d_s=2(n-1)/n$.

	We suppose that while being quite elegant the median success rule has a potential drawback that we will demonstrate on an example. Let us suppose that fitness values (to be minimized) of the previous population are say $\vc{f}_{t-1}=[2.1, 3.1, 4.1, 5.1, 6.1, 7.1, 8.1]$ while the fitness values of the current population are $\vc{f}_{t}=[1, 2, 3, 4, 5, 6, 7]$. According to the median success rule if $j$ is chosen as, e.g., 3, the number of successful individual (with fitness values better than or equal to $\vc{f}_{t-1}(3)=4.1$) is 4 (as $\vc{f}_t(1)=1, \vc{f}_t(2)=2, \vc{f}_t(3)=3$ and $\vc{f}_t(4)=4$). The computed value of $K_{succ}$ is then will be used to adapt the step-size.  However, its computation does not take into account the values of $\vc{f}_{t-1}{(i)}$ for $1\leq i<j$ and even if all such $\vc{f}_{t-1}{(i)}$ are better than the best solution $\vc{f}_t(1)$, this information will not be taken into account.
	
	This potential drawback is not the drawback in a sense that the \textit{median} success rule was designed in this way. However, we suppose that the information omitted in the median success rule can be useful since it can provide a better estimate whether and by how much the new population is more successful than the previous one. 
	
	In this paper, we introduce \textit{the population success rule} (PSR) for step-size adaptation for non-elitist multi-recombinant evolution strategies. To estimate the success of the current population we combine fitness function values from the previous and current population into a mixed set 
	
	\begin{equation} 
	\label{Fmix}
  \vc{f}_{mix} \leftarrow \vc{f}_{t-1} \cup \vc{f}_{t}
  \end{equation}
	
	Then, we rank all individual in the mixed set to define two sets $\vc{r}_{t-1}$ and $\vc{r}_{t}$ containing ranks of individuals of the previous and current populations ranked in the mixed set. 
	
	We compute a normalized success measurement
	
	\begin{equation} 
	\label{ZPSR}
  z_{PSR} \leftarrow \frac{\sum_{i=1}^{\lambda} \vc{r}_{t}(i) - \vc{r}_{t-1}(i)}{\lambda^2} - z^{*},
  \end{equation}
	where $z^*$ is a target success ratio. 
	The step-size can be adapted as in (\ref{MruleSigma}).
	
	The proposed \textit{population success rule} takes into account all fitness function values from the previous and current generation. This success rule seems to represent a more general case of the 1/5th-rule which can be obtained when $\lambda=1$.

\subsection{The Algorithm}

In the previous subsection we introduced all necessary components of the ($\mu/\mu_{w},\lambda$)-LM-CMA-ES outlined in \textbf{Algorithm} \ref{LMCMA}. 
The algorithm represents a computationally efficient limited memory version of CMA-ES, where the Cholesky factor and its inverse are reconstructed from a set of stored direction vectors (lines \ref{LMCMAsampling2} and \ref{LMCMAEvoPathUpdate}). The mutation step-size is adapted using the population success rule (lines \ref{SuccRule1} - \ref{LMCMAUpdate}). The algorithm memory and time complexity scales as $O(mn)$.

\begin{algorithm}[tb!]
\caption{The ($\mu/\mu_{w},\lambda$)-LM-CMA-ES}
\label{LMCMA}
\begin{algorithmic}[1]
\STATE{\textbf{given} $n \in \mathbb{N}_+$, $\lambda = 4 + \lfloor 3 \mstr{ln} \, n  \rfloor $, $\mu =  \lfloor \lambda/2   \rfloor $, 
											$w_i = \frac{ \mstr{ln}(\mu + 1) - \mstr{ln}(i)}{\mu \mstr{ln}(\mu + 1) - \sum_{j=1}^{\mu} \mstr{ln}(j)}; i=1 \ldots \mu$,
											$\mu_w = \frac{1}{\sum^{\mu}_{i=1} w^2_i}$,
											$c_{\sigma} = 0.3$,      
											$d_{\sigma} = 1$, $m = 4 + \lfloor 3 \mstr{ln} \, n  \rfloor $, $N_{steps} = m$,
											$c_c = \frac{1}{m}$, $c_1 = \frac{1}{10ln(n+1)}$
											} \label{LMCMAGiven}
\STATE{\textbf{initialize} $\vc{m}^{t=0} \in \R^{\dd}, \sigma^{t=0} > 0,  \vc{p}^{t=0}_{c} = \ma{0}, s \leftarrow 0 , t \leftarrow 0 $}
\REPEAT
  \FOR{$k = 1,\ldots,\lambda$} \label{LMCMAGenerateBegin}
		\STATE{ $\vc{z}_k = {{\mathcal N}  \hspace{-0.13em}\left({\ma{0},\I}\right)} $} \label{LMCMAsampling1}
			\STATE{ $\vc{x}_k = \vc{m}^t + \sigma^{t} Az( \vc{z}_k ) $} \label{LMCMAsampling2}
			\STATE{ $\vc{f}^t_k = f(\vc{x}_k)$} \label{LMCMAGenerateEnd}
  \ENDFOR
	\STATE{ $ \vc{m}^{t+1} \leftarrow \sum_{i=1}^{\mu} \vc{w}_i \vc{x}_{i:\lambda} \;$}  \label{LMCMAComputeNewMean}
	\STATE{ $ \vc{p}^{t+1}_{c} \leftarrow (1 - c_{c}) \vc{p}^{t}_{c} + \sqrt{c_{c}(2-c_{c})} \sqrt{\mu_w} (\vc{m}^{t+1} - \vc{m}^t) / \sigma$} \label{LMCMAEvoPathUpdate}
	\STATE{$ \vc{v} \leftarrow Ainvz(\vc{p}^{t+1}_c)  $} \label{LMCMAVecUpdate}
	\STATE{$ j_{cur} \leftarrow UpdateSet()$}
	\STATE{$ \vc{V}_{(j_{cur},:)} \leftarrow \vc{v};  \vc{P}_{(j_{cur},:)} \leftarrow \vc{p}^{t+1}_c$}
	\STATE{$ \vc{b}_{j_{cur}} \leftarrow \frac{\sqrt{1 - c_1}}{{\left\| \vc{v}^t \right\|}^2} \left( \sqrt{1 + \frac{c_1}{1 - c_1} {{\left\| \vc{v}^t \right\|}^2}} -1 \right) $} \label{LMCMAUpdate}
	\STATE{$ \vc{d}_{j_{cur}} =
	\frac{1}{\sqrt{1 - c_1}{\left\| \vc{v}^t \right\|}^2} 
	\left( 1- \frac{1}{\sqrt{1 + \frac{c_1}{1 - c_1} {{\left\| \vc{v}^t \right\|}^2}}} \right), 
	$}						\label{LMCMAdvec}
	\STATE{ $ \vc{r}^{t}, \vc{r}^{t-1} \leftarrow$ Ranks of $\vc{f}^{\,t}$ and $\vc{f}^{\,t-1}$ in $\vc{f}^{\,t} \cup \vc{f}^{\,t-1}$} \label{SuccRule1}
	\STATE{$ z_{PSR} \leftarrow \frac{\sum_{i=1}^{\lambda} \vc{r}^{t}(i) - \vc{r}^{t-1}(i)}{\lambda^2} - z^{*} $}
	\STATE{ $ s \leftarrow (1 - c_{\sigma})s + c_{\sigma}z_{PSR} $}
	\STATE{ $ \sigma^{t+1} \leftarrow \sigma^{t} \mstr{exp}
	          ( s / d_{\sigma} ) $} \label{LMCMAsigma}
  \STATE{ $ t \leftarrow t + 1$}
\UNTIL{ \textit{stopping criterion is met} }
\end{algorithmic}
\end{algorithm}

\section{Simulation Results}
\label{expe}

In this section, we perform a set of numerical experiments to assess the performance of the proposed LM-CMA-ES on large  scale optimization problem with $n=128,256,\ldots,8096$. We investigate the performance on three basic problems: Sphere function $f_{Sphere}(\vc{x}) = \sum_{i=1}^n \vc{x}^{2}_{i}$, separable Ellipsoid function $f_{Elli}(\vc{x})=\sum_{i=1}^n 10^{6\frac{i-1}{n-1}} \vc{x}^{2}_{i}$ and its rotated version $f_{ElliRot}(\vc{x})=f_{Elli}(\vc{Q}\vc{x})$, where $\vc{Q}$ is an orthogonal $n \times n$ matrix with each column vector $\vc{q}_i$ being a uniformly distributed unit vector implementing an angle-preserving transformation \cite{ros2008simple}.

\subsection{Experimental Setting}

For the sake of reproducibility, the MATLAB/C++ source code of all tested algorithms is available at \\ \url{https://sites.google.com/site/lmcmaeses/}. 

In the order to estimate the performance of ($\mu/\mu_{w},\lambda$)-LM-CMA-ES, we compare it with ($\mu/\mu_{w},\lambda$)-Cholesky-CMA-ES and ($\mu/\mu_{w},\lambda$)-Sep-CMA-ES. We use the default parameters for Cholesky-CMA-ES and Sep-CMA-ES as given in \cite{2009Suttorp11CMA} and \cite{ros2008simple}, respectively. The parameters of LM-CMA-ES are given in \textbf{Algorithm} \ref{LMCMA}. For all problems, the mean $\vc{m}^{t=0}$ is initialized in the range $[-5,5]^n$, the population is sampled with initial step-size $\sigma^{t=0}=5$ and using the same seed per run. Note that in all cases we use the default population size $\lambda=4 + \lfloor 3 \mstr{ln} \, n  \rfloor$.
\newpage

\subsection{Memory and Computational Complexity of LM-CMA-ES}

The LM-CMA-ES has $O(mn)$ memory complexity and more specifically stores $\vc{Q} \in \R^{m \times n}$, $\vc{V} \in \R^{m \times n}$ and $\lambda$ solution vectors $\vc{x}_i$. For large $n$ and $m=\lambda_{default}=\lambda=4 + \lfloor 3 \mstr{ln} \, n  \rfloor$ used in this paper, the algorithm stores approximately $3mn$ real-valued parameters. If a real-valued parameter requires $8$ bytes of memory, then for $n=8192$ the LM-CMA-ES will require 5.8 megabytes while the original CMA-ES would start to reach its limit by requiring 1 gigabyte of memory. Using the same amount of memory (more specifically, 1.03 gigabyte), the LM-CMA-ES will able to optimize a 1 million dimensional problem. Indeed, by taking $m=1$ even less memory would be needed but the latter possibility makes sense only if the performance stays at a reasonable level. 

\begin{figure}[t]
\centerline{ 
	\includegraphics[width=3.85truein]{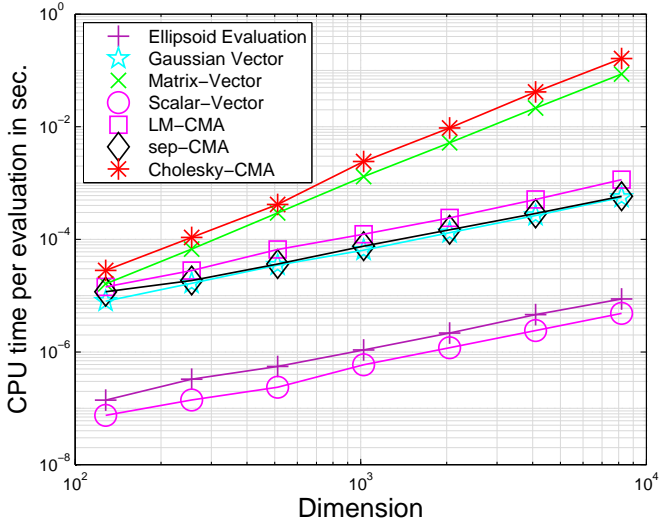} 
}
\caption{\label{fig:timing} Timing results of LM-CMA-ES on the separable Ellipsoid compared to sep-CMA-ES and Cholesky-CMA-ES. 
 The results were computed using at most $10^5$ function evaluations for sep-CMA-ES and LM-CMA-ES and using at most $10^4$ for Cholesky-CMA-ES.  
}
\end{figure}

Figure \ref{fig:timing} shows how fast CPU time per evaluation scales for different operations (measured on a 2.0 GHz processor). Scalar-vector multiplication of a vector with $n$ variables scales linearly with ca. $6\cdot{10}^{-10}n$ seconds, evaluation of the separable Ellipsoid is twice more expensive if a temporary data is used.
 Sampling of $n$ normally distributed variables scales as ca. 100 vectors-scalar multiplications. As can be seen in Figure \ref{fig:timing}, sampling of $\vc{z}_k$ dominates the computational overhead of sep-CMA already after $n=128$. The LM-CMA-ES scales almost linearly for $n\geq1024$ as ca. $1.3\cdot{10}^{-7}n$ or ca. 200 scalar-vector multiplications. Matrix-vector multiplication scale quadratically with $n$ and Cholesky-CMA-ES scales as ca. 1.5-2 matrix-vector multiplications. 

Practically, the LM-CMA-ES is about 40 times faster (in terms of its internal computation cost per function evaluation) for $n=2048$ and about 140 times faster for $n=8192$ than Cholesky-CMA-ES. The LM-CMA-ES is only about 2 times slower than sep-CMA-ES, whose cost is dominated by sampling from normal distribution.

The computation cost of CMA-ES with full covariance matrix learning limits its applicability for $n\gg100$ and makes it intractable because of memory for $n>10000$.

\begin{figure}[t]
\centerline{ 
	\includegraphics[width=3.85truein]{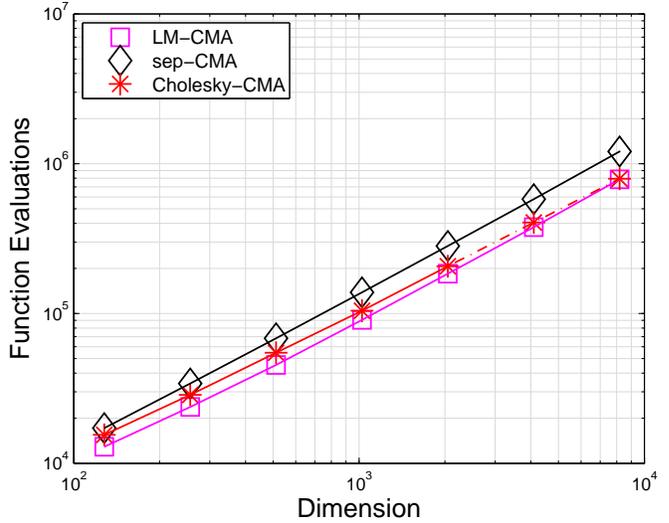}
}
\caption{\label{fig:sphere} Results of LM-CMA-ES on the Sphere function compared to sep-CMA-ES and Cholesky-CMA-ES. Lines show the median of 11 runs for different problem dimensions to reach the target fitness value of $10^{-10}$. The dotted line is an extrapolation.
}
\end{figure}

\subsection{Performance on Sphere and Ellipsoid}

The Sphere function is often viewed in Evolutionary Computation to be the first function to look at when benchmarking evolutionary algorithms. Figure \ref{fig:sphere} demonstrates a comparable performance of LM-CMA-ES with population success rule, sep-CMA-ES with CSA and Cholesky-CMA-ES with CSA. 
It should be further studied what is the effect of the target population success rate (set to 
$z^*=0.25$) whose value was chosen the same for all experiments in order to obtain a reasonable performance on Ellipsoid functions.

\begin{figure*}[t]
\centerline{ 
	\includegraphics[width=3.85truein]{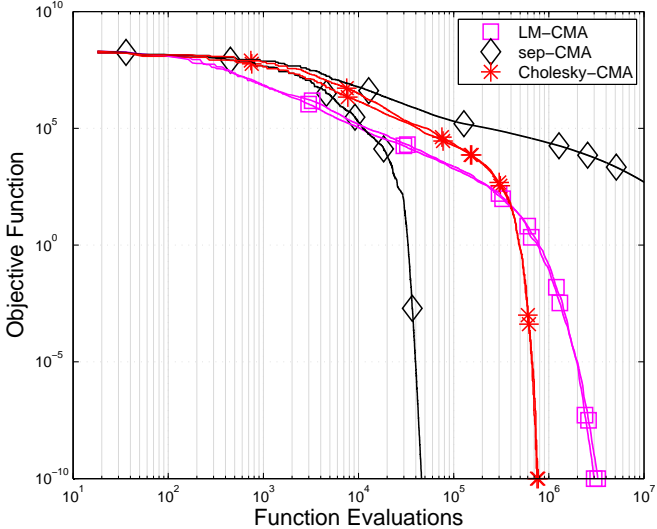}
	\includegraphics[width=3.85truein]{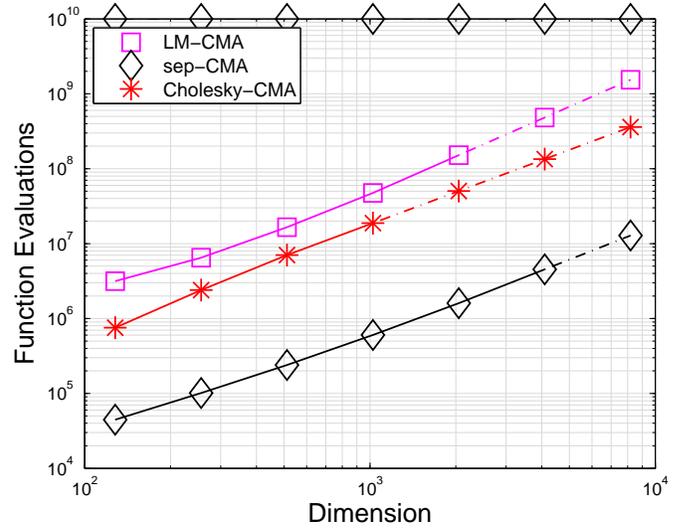}
}
\caption{\label{fig:elli} \textbf{Left}: Convergence plots of LM-CMA-ES, sep-CMA-ES and Cholesky-CMA-ES
 on 128-dimensional axis-parallel and rotated Ellipsoid functions.
 \textbf{Right}: The median of 11 runs on separable Ellipsoid function 
for different problem dimensions. The dotted lines correspond to extrapolated results by preserving the same scaling as between the last two actual estimations. 
}
\end{figure*}

Figure \ref{fig:elli}-Left shows that both LM-CMA-ES and Cholesky-CMA-ES are rotationally invariant and therefore they optimization runs (one per function) are almost coincide (within the algorithm). The sep-CMA-ES is not rotationally invariant and therefore it performs better on the separable Ellipsoid than on its rotated version where the exploitation of the separability is not that useful. 
Importantly, the LM-CMA-ES often outperforms the Cholesky-CMA-ES in the beginning of optimization, while
 the adaptation of the full covariance matrix makes Cholesky-CMA-ES faster at later stages. 
Figure \ref{fig:elli}-Right shows that the loss of performance of LM-CMA-ES compared to Cholesky-CMA-ES is in order of a factor of 3-4 given that for $n=2048$ the LM-CMA-ES uses only $m=26$ direction vectors. It is important to keep in mind that for $n>10000$ the Cholesky-CMA-ES becomes intractable both due to its memory and computational complexity. Then, the sep-CMA-ES becomes an alternative, however, it does not learn dependencies and might be therefore inefficient (see Figure \ref{fig:elli}-Left). 

We discussed several large scale ESs in this paper: L-CMA-ES \cite{knight2007reducing} and R1-NES \cite{sun2011linear}. 
We compared the LM-CMA-ES indirectly by analyzing the results from \cite{knight2007reducing} and \cite{sun2011linear}. It takes about 6000 seconds for L-CMA-ES to solve 200-dimensional Ellipsoid after about $7e+6$ function evaluations with $m=\sqrt{n}=14$ and 4000 seconds after $4e+6$ evaluations with $m=n/2=100$. The LM-CMA-ES solves the same problem after about $125$ seconds and $5.3e+6$ function evaluations with $m=19$. The performance is comparable while the LM-CMA-ES is about $32-48$ times faster that is unlikely to be only due to a different processor or implementation used. The L-CMA-ES has $O(m^2n)$ computational complexity and therefore it is in order of $m$ times computationally slower than LM-CMA-ES.

The R1-NES algorithm performs well on non-separable problems but tends to fail on problems where the learning of multiple principal components is essential, e.g., it fails on moderate dimensional rotated Ellipsoid function \cite{sun2011linear}. On Rosenbrock function the LM-CMA-ES is about 5 times faster (not shown) in terms of number of function evaluations for $n=256,512$. The R1-NES also samples from the normal distribution, and therefore the lower bound of its computational complexity is predefined (see Figure \ref{fig:timing}). 

\begin{figure}[t]
\centerline{ 
	\includegraphics[width=3.85truein]{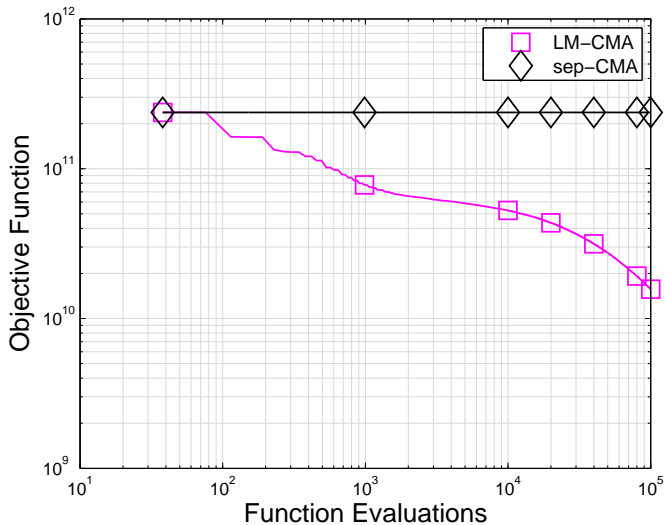}
}
\caption{\label{fig:Elli100k} LM-CMA-ES and sep-CMA-ES on separable 100,000-dimensional Ellipsoid problem. The sep-CMA-ES divergences after the first generation (the best fitness is shown). Note that the LM-CMA-ES is rotationally invariant, therefore a similar performance is expected on 100,000-dimensional rotated Ellipsoid.  
}
\end{figure}

We performed an experiment on 100,000-dimensional separable Ellipsoid problems for 100,000 function evaluations (i.e., $n$ evaluations). The original CMA-ES and Cholesky-CMA-ES cannot be applied due to memory requirements. The applicability of L-CMA-ES is also limited due to its $O(m^2n)$ computational complexity. The results for sep-CMA-ES specifically designed for large scale optimization and the proposed LM-CMA-ES are shown in Figure \ref{fig:Elli100k}. While the LM-CMA-ES gradually improves the fitness similarly as in Figure \ref{fig:elli}-Left, the sep-CMA-ES does not improve it because it diverges from the very first iterations. To investigate whether it is a mistake in our implementation, we launched the same experiment using the sep-CMA-ES author's MATLAB implementation where
 the divergence was also observed. 

It should be noted that the separable Ellipsoid can be easily solved by various Evolutionary Algorithms which 
implicitly or explicitly exploit its separability, our purpose of its usage is to investigate how the LM-CMA-ES performs on problems with high dependencies between variables. Given that the LM-CMA-ES is rotationally invariant, its performance on both separable and non-separable problems is comparable, but the former is cheaper to compute.

%%%%%%%%%%%%%%%%%%%%%%%%%%%%%%%%%%%%%
\section{Discussion and Conclusion}
%%%%%%%%%%%%%%%%%%%%%%%%%%%%%%%%%%%%%
\label{conclusion}

This paper presents a new approach to efficiently store and exploit the information about dependencies between decision variables of large scale optimization problems. 
 It allows to reconstruct the Cholesky factor and its inverse using $m\ll n$ direction vectors that turns out to be sufficient to obtain good performance on large scale problems with highly-depended variables. 
 The implementation of this approach in the LM-CMA-ES algorithm makes it possible to optimize a 1 million dimensional problem while learning dependencies between variables at a cost of about 0.1 second per function evaluation on an ordinary machine. Indeed, one should not plan to easily find a global optimum in such a huge search space, but some local optimization/tuning seems reasonable, e.g., in Machine Learning problems.

The proposed LM-CMA-ES algorithm is based on the \textit{population success rule} which looks promising and requires further theoretical and empirical investigations. It should be studied as well whether it can be claimed to represent a general case of the 1/5th success rule. 
More experiments are required to investigate whether and when the lack of rank-$\mu$ update is a limitation.

All parameters chosen for the algorithm were tuned only moderately and \textit{specifically} for large $n$ and might require a significant revision to address a wider set of optimization problems commonly used for EAs. However, we suppose that the performance on the Ellipsoid 
 function is already worth a closer scientific investigation. We envision that several directions may further improve the algorithm: i) adaptation of $m$ within a fixed range, the impact of $m$ itself should be studied as well, ii) since the population success rule does not make any assumptions about the sampling distribution, the Gaussian sampling can be removed that would further speed-up the algorithm (e.g., to replace CSA by PSR in CMA-ES). 

The speculations about a possibility of having CMA-ES like evolutionary processes going on in nature often end up around a hypothesis that there is no such a thing  in natural evolution as a full covariance matrix and its update. One may suppose that only a limited number of direction vectors is stored to adjust the mutation in promising directions.

\bibliographystyle{abbrv}
\small
%\bibliography{gecco14}

\end{document}